\documentclass[11pt]{article}

\usepackage[a4paper,margin=2.5cm]{geometry}
\usepackage{amsmath,amssymb}
\usepackage{bm}
\usepackage{booktabs}
\usepackage[section]{placeins}

\usepackage{graphicx}
\usepackage[labelfont=normalfont,justification=centering]{subcaption}
\DeclareCaptionLabelFormat{italicparens}{(\textit{#2})}
\usepackage[numbers]{natbib}
\usepackage{hyperref}
\usepackage{orcidlink}
\usepackage{adjustbox}

\newcommand{\ntrainsims}{32}
\newcommand{\nsplits}{10}
\newcommand{\nsplitsThreeD}{5}
\newcommand{\pairRecoveryZero}{$3.0 \times 10^{-9}$}
\newcommand{\pairRecoveryOne}{$1.9 \times 10^{-8}$}
\newcommand{\pairRecoveryTwo}{$2.4 \times 10^{-8}$}
\newcommand{\pairRecoveryThree}{$1.5 \times 10^{-8}$}
\newcommand{\recoveryZero}{$5.8 \times 10^{-7}$}
\newcommand{\recoveryOne}{$8.1 \times 10^{-6}$}
\newcommand{\recoveryTwo}{$1.8 \times 10^{-5}$}
\newcommand{\recoveryThree}{$1.3 \times 10^{-5}$}
\newcommand{\mixedRecoveryZero}{$9.0 \times 10^{-5}$}
\newcommand{\mixedRecoveryOne}{$9.2 \times 10^{-4}$}
\newcommand{\mixedRecoveryTwo}{$2.2 \times 10^{-3}$}
\newcommand{\mixedRecoveryThree}{$1.9 \times 10^{-3}$}
\newcommand{\mixedThreeDRecoveryZero}{$1.8 \times 10^{-4}$}
\newcommand{\mixedThreeDRecoveryOne}{$1.9 \times 10^{-3}$}
\newcommand{\mixedThreeDRecoveryTwo}{$3.7 \times 10^{-3}$}
\newcommand{\mixedThreeDRecoveryThree}{$3.6 \times 10^{-3}$}

\newcommand{\R}{\mathbb{R}}
\newcommand{\C}{\mathbb{C}}
\newcommand{\bv}{\bm{v}}
\newcommand{\br}{\bm{r}}
\newcommand{\bu}{\bm{u}}
\newcommand{\ba}{\bm{a}}
\newcommand{\msl}{{\{\hspace{-0.75ex}\{}}
\newcommand{\msr}{{\}\hspace{-0.75ex}\}}}
\newif\ifjournal \journaltrue

\title{Interpretable many-body interaction laws for collective motion from few trajectories}
\author{Timon S. Gutleb\,\orcidlink{0000-0002-8239-2372}\\ \small School of Computer Science, University of Leeds, Leeds, UK\\
  \small \texttt{T.S.Gutleb@leeds.ac.uk}}
\date{}

\begin{document}
\maketitle

\begin{abstract}
Models of collective motion usually reduce social interactions to pairwise
forces, despite empirical evidence for many-body effects. We introduce the
directional atomic cluster expansion (dACE), which adapts the atomic cluster
expansion (ACE) of interatomic potentials to self-propelled agents. dACE provides a
systematically improvable, exactly equivariant expansion of an agent's
acceleration that splits uniquely into 1-, 2-, 3-body and higher contributions,
along with a self term for the agent's own past motion that is fitted so that it does not
absorb the interactions. On simulated collectives in two and three dimensions with
known rules, dACE recovers the interaction law body order by body order from little
data, while the neural model SocialNet, the current closest competitor, needs more data,
has much larger errors on many-body or chiral rules and is restricted to 2D. On real data sets of tracked zebrafish
and golden shiners, the laws learned by dACE agree better between disjoint halves of
the data than SocialNet's. The learned laws are interpretable, and in the shiners the
3-body term opposes the pairwise response as neighbourhoods get dense. Finally, we
show that next-step prediction accuracy mostly measures self-propulsion and noise
rather than the interactions, and suggest more robust combinations of error measures
for interaction learning.
\end{abstract}

\noindent\textbf{Keywords:} collective motion, swarms, interaction rules, many-body interactions, atomic cluster expansion (ACE), self-propelled particles, zebrafish, golden shiners

\section{Introduction}
\label{sec:intro}

Many animals move collectively, from insect swarms and fish schools to bird
flocks and ungulate herds \citep{vicsekzafeiris2012}, and robotic swarms are
engineered to behave similarly \citep{brambilla2013}.
Understanding the rules by which individuals respond to one another is a central
question: in biology, to compare species and test evolutionary
hypotheses, and in robotics, to design swarms and to defensively anticipate the behaviour of
black-box swarms whose rules are not known.

The classical way to model collective motion is to prescribe social forces
between pairs of individuals, such as short-range repulsion, alignment and
long-range attraction, and to sum or average them over a neighbourhood
\citep{reynolds1987,vicsek1995,couzin2002,dorsogna2006,cuckersmale2007}. Studies that fit such models to tracked trajectories often fix the functional forms in
advance and infer only weights, ranges and the neighbourhood model
\citep{lukeman2010,torney2018}. Pairwise models are
interpretable and cheap, but there is mounting evidence that interactions in
animal groups do not reduce to sums or averages of pairwise forces. In golden
shiner schools, the force on a fish with two neighbours contains substantial
three-body contributions beyond the average of the two pairwise forces
\citep{katz2011}, and marching locusts, the textbook case of a density-driven
alignment transition \citep{buhl2006}, respond to a virtual swarm in ways that no
alignment force captures \citep{sayin2025}. Robotic swarms, whose rules are designed rather than evolved, have even less reason to be restricted to pairwise rules. What is missing is a systematic way to
learn interaction rules of arbitrary body order from trajectory data, without
prescribing their form, and to determine how much each body order contributes.

Methods that learn interaction rules without fixing their form include
non-parametric inference of pairwise radial kernels
\citep{lu2019,lu2022}, force maps and reconstructions built on pairwise or
averaged pairwise contributions \citep{katz2011,herbertread2011},
sparse regression over prescribed libraries \citep{brunton2016}, and neural
networks, among them networks on interaction graphs \citep{cranmer2020},
neural models of pairs of fish \citep{papaspyros2024} and of influence between animals
\citep{fujii2021}, networks that design interaction rules for desired collective
patterns \citep{kim2025} and symbolic models distilled
from graph networks \citep{powers2025}. Of the learned models tested on animal
groups, only the deep attention network of \citet{heras2019}, developed to predict the turns of
zebrafish in groups of 60 to 100 and released as SocialNet with an option to predict accelerations,
handles large groups and has public code, so we compare with it in 2D (as it does not support 3D). SocialNet's nonlinear aggregation mixes all orders of
interaction, so the contribution of each body order is not directly accessible,
and, as we show, it needs more data than dACE. Data, however, is scarce in this field, since extracting reliable trajectories
of many individuals remains difficult despite much work on tracking \citep{perezescudero2014,romeroferrero2019},
and three-dimensional field data require enormous effort and dedicated campaigns
\citep{cavagna2008starflag,ling2019}.

In atomic physics and computational chemistry, the analogous problem of learning many-body
interatomic potentials for interacting particles has a rich history of polynomial and body-ordered models
\citep{braams2009,musil2021}. The
atomic cluster expansion (ACE) \citep{drautz2019} made body-ordered models of
atoms efficient at high body order, as a complete, systematically improvable linear
expansion in symmetry-adapted basis functions of an atom's neighbourhood
\citep{dusson2022,bachmayr2024}. It underlies some of the most accurate interatomic
potentials \citep{kovacs2021,witt2023acepotentials} and the highly successful MACE
networks \citep{batatia2022mace}. Recently, the
self-interaction-free (pure) formulation of Ho, Gutleb and Ortner~\citep{ho2024pure} made the
coefficients of each body order separately identifiable while retaining efficiency.

In this paper we introduce the \emph{directional} atomic cluster expansion (dACE), a
pure, exactly equivariant expansion for self-propelled interacting agents in 2D and 3D with a staged self term for
their own past motion (Sections~\ref{sec:method} and \ref{sec:selfterm}). We test it on
simulated collectives with known 2-, 3- and 4-body rules and on tracked zebrafish and
golden shiners, against SocialNet (Sections~\ref{sec:results} to \ref{sec:realdata}), and
close with lessons for learning interaction laws in general and directions for future
work (Section~\ref{sec:limitations}). We also release an open-source Julia implementation, \texttt{dACE.jl} \citep{gutleb2026dacejl}.

\section{Directional atomic cluster expansion}
\label{sec:method}

In atomic physics and chemistry the underlying laws are believed to be pairwise, yet
many-body expansions substantially improved the learned models of the effective
interactions between atoms in complex structures. For the rules of biological
collective motion there is little reason to expect them to be pairwise, since the
responses of animals pass through perception and time-sensitive decision-making. A many-body rule
need not be the complicated nonlinear combination of neighbours it is often taken for
\citep{bialek2012}, since a response to neighbours in loose groupings of twos and threes is a rule of higher body order.

\subsection{The atomic cluster expansion}
\label{sec:ace}

\paragraph{Body-ordered expansion.}
ACE approximates functions $V(X)$ of a multiset $X = \msl x_j\msr_{j=1}^J$ of
$J$ particles $x_j \in \Omega$, typically the neighbours of a central atom with
positions relative to it and chemical species, $x_j = (\br_j, Z_j)$
\citep{drautz2019,ho2024pure}. The first step is a truncated many-body
expansion,
\[
  V(X) \approx \sum_{N=0}^{\mathcal{N}} \sum_{j_1 < \dots < j_N} v_N(x_{j_1}, \dots, x_{j_N}),
\]
in which each permutation-invariant component $v_N$ is expanded in a tensor
product of one-particle basis functions $\phi_k : \Omega \to \C$. Collecting
coefficients gives the canonical, or pure, cluster expansion
\[
  V(X) \approx \sum_{\bm{k}} c_{\bm{k}}\, \mathcal{A}_{\bm{k}}(X),
  \qquad
  \mathcal{A}_{\bm{k}}(X) = \sum_{j_1 \neq \dots \neq j_N} \prod_{t=1}^{N} \phi_{k_t}(x_{j_t}),
\]
over ordered tuples $\bm{k} = (k_1 \le \dots \le k_N)$ of correlation order $N$
(body order $N + 1$, counting the central particle). We count \emph{body orders}
throughout this paper, as is usual for collectives: the 1-body term depends on the
agent alone, the 2-body terms are the pairwise interactions, and so on. Evaluated naively, the sum over distinct
tuples costs $O(J^N)$.

\paragraph{Self-interaction and purification.}
\citet{drautz2019} replaced the sum over distinct tuples by the sum over all
tuples, which factorises into pooled one-particle features,
\[
  A_{\bm{k}}(X) = \prod_{t=1}^{N} A_{k_t}(X), \qquad A_k(X) = \sum_{j} \phi_k(x_j),
\]
at a cost of $O(J)$ per feature $A_k$ plus $O(N)$ per product. The price of this efficiency gain is the inclusion of tuples with repeated
indices, the ``self-interactions''. Ho, Gutleb and
Ortner~\citep{ho2024pure} showed that the pure basis can be recovered at
comparable cost when pointwise products of one-particle functions can be
linearised, $\phi_k\phi_{k'} = \sum_{k''} \pi^{kk'}_{k''} \phi_{k''}$, as is the
case for polynomial and trigonometric bases. Then $\mathcal{A} = \mathcal{P} A$
for a sparse, precomputed purification operator $\mathcal{P}$, and the pure basis
separates the contributions of the body orders. A pure basis function of correlation
order $N$ vanishes on every environment with fewer than $N$ neighbours, so
the body-order decomposition is unique and fixed by the data.

\paragraph{Symmetry.}
Physical potentials are often invariant under transformations such as rotations
and reflections of the whole configuration, and the forces they generate are then
equivariant. ACE imposes such symmetries exactly. The
one-particle functions $\phi_{nlm}(\br, Z) = R_{nZ}(|\br|)\, Y_l^m(\hat\br)$ transform
under rotations by known matrices, and their products are coupled to invariant or
equivariant combinations, $\mathcal{B} = \mathcal{C}\mathcal{A}$, with generalised
Clebsch--Gordan coefficients $\mathcal{C}$ computed once
\citep{drautz2019,witt2023acepotentials}. Symmetrisation and purification are
both sparse linear maps and compose, $\mathcal{B} = \mathcal{C}\mathcal{P}A$
\citep{ho2024pure}. In two dimensions, the spherical harmonics become circular
harmonics $e^{im\theta}$ and the coupling reduces to a selection rule: a
product carries the total charge $\sum_t m_t$, and the invariant (equivariant)
products are those with total charge 0 (that of the output).

\subsection{From ACE to dACE}
\label{sec:dace}

At least two important things separate agents from atoms in terms of general modelling. (i) Animals and robots can perceive in which
direction their neighbours move and respond relative to their own heading, so the
expansion has to include velocities, and the only symmetry left is a joint rotation
of positions and velocities. (ii) Atoms do not propel themselves, whereas much of
an animal's or a drone's motion is its own. A model of collective motion therefore needs a self model
next to the interactions that does not absorb them, despite an agent's own past
motion being correlated with its neighbours'. The same correlation makes next-step
prediction accuracy a misleading measure of interaction models
(Section~\ref{sec:selfresults}). We address (i) below and (ii) in
Section~\ref{sec:selfterm}.

\begin{figure}[tbp]
  \centering
  \begin{subfigure}[b]{0.34\linewidth}\centering\includegraphics[width=\linewidth]{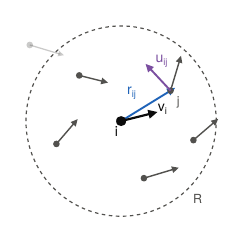}\caption{}\end{subfigure}\hfill
  \begin{subfigure}[b]{0.62\linewidth}\centering\includegraphics[width=\linewidth]{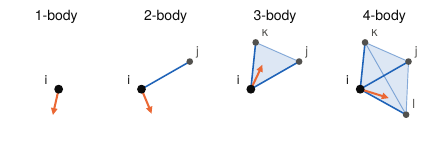}\caption{}\end{subfigure}
  \caption{dACE. (a) The environment of agent $i$: its velocity $\bv_i$ and, for each
    neighbour within $R$, the displacement $\br_{ij}$ and relative velocity $\bu_{ij}$.
    (b) The force on $i$, $\bm F = \bm F^{(1)} + \bm F^{(2)} + \cdots$, with $\bm F^{(N+1)}$ summed over
    all groups of $N$ neighbours.}
  \label{fig:schematic}
\end{figure}

\paragraph{Local environments.}
Agent $i$ at time $t$ has position $\bm{x}_i$ and velocity $\bv_i$ in $\R^d$,
$d \in \{2, 3\}$. Its environment consists of its own velocity $\bv_i$ and the
multiset
\[
  \mathcal{X}_i = \msl (\br_{ij}, \bu_{ij}) : 0 < |\br_{ij}| < R \msr,
  \qquad \br_{ij} = \bm{x}_j - \bm{x}_i, \quad \bu_{ij} = \bv_j - \bv_i ,
\]
as sketched in Figure~\ref{fig:schematic}a. We learn the acceleration $\ba_i = \bm{F}(\bv_i, \mathcal{X}_i)$. Compared with atomistic ACE, each
neighbour now carries two vectors. We expand the acceleration directly in equivariant vector-valued basis functions, as ACE
expands vectorial and tensorial properties \citep{drautz2020}. We assume no potential
and construct none, and the acceleration need not be a gradient, since alignment and self-propulsion are not
conservative. The symmetries relevant for dACE are permutations of the neighbours and joint rotations of all positions and
velocities, $\bm{F}(Q\bv_i, Q\mathcal{X}_i) = Q\bm{F}(\bv_i, \mathcal{X}_i)$.
Positions and velocities must be rotated together, since agents respond to
their neighbours' directions of motion relative to their own heading. This joint action of one rotation on pairs of vectors, one pair per neighbour, has no current counterpart in atomistic ACE and is of interest in its own right.

\paragraph{Pure body-ordered dACE basis.}
Given one-particle functions $\phi_k(\br, \bu)$ and centre functions
$\psi(\bv)$ (specified for $d = 2, 3$ below), the dACE basis functions
(Figure~\ref{fig:schematic}b) are,
for a multi-index $\bm{k} = (k_1 \le \dots \le k_N)$,
\[
  B_{\psi\bm{k}}(\bv_i, \mathcal{X}_i) = \psi(\bv_i)\, \mathcal{A}_{\bm{k}}(\mathcal{X}_i)
  = \psi(\bv_i) \sum_{j_1 \neq \dots \neq j_N} \prod_{t=1}^N \phi_{k_t}(\br_{ij_t}, \bu_{ij_t}).
\]
The sum over distinct neighbours expands by inclusion--exclusion into products of pooled
features, with $\mathcal{T}_N$ the set partitions of $\{1, \dots, N\}$,
\[
  \sum_{j_1 \neq \dots \neq j_N} \prod_{t} \phi_{k_t}(\br_{ij_t}, \bu_{ij_t})
  = \sum_{\tau \in \mathcal{T}_N} \mu(\tau) \prod_{I \in \tau} A_{\bm{k}_I},
  \qquad
  A_{\bm{k}_I} = \sum_j \prod_{t \in I} \phi_{k_t}(\br_{ij}, \bu_{ij}),
\]
with $\mu(\tau) = \prod_{I\in\tau} (-1)^{|I|-1}(|I|-1)!$. The block sums
$A_{\bm{k}_I}$ are shared between basis functions and computed once per
environment. We compute the multi-element block sums by purification: the
product of the one-particle functions in a block is linearised exactly in a
basis of pooled features (given below), and the block sums follow by a sparse
precomputed linear map analogous to that of Ho, Gutleb and
Ortner~\citep{ho2024pure}. All one-particle functions carry the smooth cutoff
$f_c(\rho) = (1-\rho^2/R^2)^2$, which is not closed under multiplication, so we pool features separately for each power $h = |I|$ of $f_c$. Neighbours
then enter only through the pooled features, at a cost $O(J)$ per feature, and we obtain the pure basis at a cost comparable to the self-interacting one.
As in ACE \citep{drautz2019}, we take the radial dependence through polynomials of a transformed distance, here Legendre polynomials $P_n(y(\rho))$ of $y(\rho) = 1 - 2(1-e^{-\rho/\ell})/(1-e^{-R/\ell})$, which resolves short distances,
where interactions are strongest, more finely than a polynomial in $\rho$, and scale
velocities by a typical speed $v_s$, $\tilde\bu = \bu / v_s$, $\tilde\bv = \bv / v_s$. We use
$\ell = R/2$.

The speed polynomials $P_l(2|\tilde u|^2 - 1)$ are orthogonal on $|\tilde u| \le 1$.
When speeds are heavy-tailed, as for animals that alternate bursts and glides, a
scale $v_s$ that covers the fast tail leaves most observations at small
$|\tilde u|$, where the polynomials resolve little. Our implementation therefore also includes an
optional saturating map
\begin{equation}
  \tilde{\bm w} = \tanh(|\bm w|/v_s)\, \hat{\bm w},
  \label{eq:speedmap}
\end{equation}
which places every speed in the unit ball, resolves speeds below $v_s$ finely and
saturates above. It keeps directions, so it commutes with rotations and
reflections, and velocity factors remain smooth at $\bm w = 0$, since $\tanh(t)/t$
is an even analytic function. The same map, with a scale $a_s$, applies to the past accelerations of
the self term (Section~\ref{sec:selfterm}). Purification uses only identities in the
scaled variables and is unchanged. We use the linear scaling throughout, except where stated.

\subsection{Two dimensions}
\label{sec:method2d}

\paragraph{Complex encoding and rotations.}
Rotation equivariance in 2D reduces to the charge bookkeeping of Section~\ref{sec:ace},
which we now make precise. We identify $\R^2 \cong \C$, $z = x + iy$. A joint rotation by $\theta$ of all
positions and velocities acts as $z \mapsto e^{i\theta} z$. A function $f$ has
charge $q \in \mathbb{Z}$ if $f \mapsto e^{iq\theta} f$. Writing
$z^{m} := \bar z^{|m|}$ for $m < 0$, the factor $z^m$ has charge $m$, and charges
add under multiplication. The acceleration $a_x + i a_y$ has charge $1$.

\paragraph{One-particle and centre bases.}
With $\rho = |\br|$, $\hat r = (r_x + i r_y)/\rho$ and $\tilde u = \tilde u_x + i \tilde u_y$,
\begin{align*}
  \phi_{n l m_r m_u}(\br, \bu) &= f_c(\rho)\, P_n(y(\rho))\, P_l(2|\tilde u|^2 - 1)\, \hat r^{m_r}\, \tilde u^{m_u}, \\
  \psi_{p m}(\bv) &= P_p(2|\tilde v|^2 - 1)\, \tilde v^{m},
\end{align*}
with charges $q(\phi) = m_r + m_u$ and $q(\psi) = m$. Velocity factors are polynomials in $\tilde\bu$ and $\tilde\bv$, so they are smooth at
$\bu = 0$ and $\bv = 0$ under either scaling.

\paragraph{Equivariance.}
We keep only $(\psi, \bm{k})$ with total charge $q(\psi) + \sum_t q(\phi_{k_t}) = 1$, and a
bound on the total degree $\deg(\psi) + \sum_t \deg(\phi_{k_t})$ per body order,
where $\deg(\phi_{n l m_r m_u}) = n + 2l + |m_r| + |m_u|$ and $\deg(\psi_{pm}) =
2p + |m|$ count the polynomial degree in $y$, $\tilde u$ and $\tilde v$. Then every basis function, and hence
$\bm F = \sum c_{\psi\bm{k}} B_{\psi\bm{k}}$ for any complex coefficients, is exactly
equivariant under joint rotations of positions and velocities, and no Clebsch--Gordan coefficients are needed.

\paragraph{Reflections.}
A reflection conjugates all positions and velocities, and maps every basis
function to its complex conjugate, $B_{\psi\bm{k}}(\sigma \bv_i, \sigma \mathcal{X}_i) =
\overline{B_{\psi\bm{k}}(\bv_i, \mathcal{X}_i)}$. Hence $\bm F$ is also equivariant under
reflections if and only if all coefficients are real. Real coefficients halve the
number of parameters, and we use them for every mirror-symmetric (achiral) collective. Complex
coefficients allow chiral rules to be learned, but on achiral collectives they
incur an avoidable loss of accuracy and efficiency.

\paragraph{Purification.}
The pooled features are
\[
  \tilde A_{h\nu L M_r M_u} = \sum_j f_c(\rho_{ij})^h\, P_\nu(y(\rho_{ij}))\,
  P_L(2|\tilde u_{ij}|^2 - 1)\, \hat r_{ij}^{M_r}\, \tilde u_{ij}^{M_u},
\]
using Legendre linearisation in $y$ and $|\tilde u|^2$, $\hat r^{m}\hat r^{m'} =
\hat r^{m+m'}$, and $\tilde u^{m}\tilde u^{m'} = |\tilde u|^{|m| + |m'| - |m + m'|}\,\tilde u^{m+m'}$.

\subsection{Three dimensions}
\label{sec:method3d}

\paragraph{Heading frame.}
In 3D each neighbour still carries two directions, $\hat\br$ and $\hat\bu$, but on
the sphere their harmonics no longer combine by adding charges, and full
symmetrisation would couple them with generalised Clebsch--Gordan coefficients. We use instead that self-propelled agents have a heading: the focal velocity defines a
right-handed orthonormal frame $(\bm e_1, \bm e_2, \bm e_3)$ with $\bm e_3 =
\bv_i/|\bv_i|$, in which every vector $\bm w$ is described by its perpendicular
part $w_\perp = \bm w \cdot \bm e_1 + i\, \bm w \cdot \bm e_2 \in \C$ and its axial
part $w_\parallel = \bm w \cdot \bm e_3$. A joint rotation $Q$ maps the frame of
$\bv_i$ to that of $Q\bv_i$ up to a rotation by some $\theta$ about $\bm e_3$,
which multiplies all perpendicular parts by $e^{i\theta}$ and leaves the axial
parts unchanged. Rotations about the heading are therefore handled by the same
charge selection rule as in 2D, and the arbitrary choice of $\bm e_1$ affects no
prediction. An agent at rest has no heading, so we keep the frame of its latest
non-zero velocity, which leaves the model defined everywhere and exactly equivariant
under joint rotations whenever $\bv_i \neq 0$.

\paragraph{One-particle and centre bases.}
With $\hat\br = (\hat r_\perp, \hat r_\parallel)$ and $\tilde\bu = (\tilde u_\perp, \tilde u_\parallel)$,
\begin{align*}
  \phi_{n a m_r l b m_u}(\br, \bu) &= f_c(\rho)\, P_n(y(\rho))\, P_a(\hat r_\parallel)\, \hat r_\perp^{m_r}\,
    P_l(2|\tilde u_\perp|^2 - 1)\, P_b(\tilde u_\parallel)\, \tilde u_\perp^{m_u}, \\
  \psi_p(\bv) &= P_p(2|\tilde v| - 1),
\end{align*}
where $\hat r_\perp^{m_r} P_a(\hat r_\parallel)$ are smooth functions on the sphere, with charges $q(\phi) = m_r + m_u$ and $q(\psi) = 0$ and degrees $n + a +
|m_r| + 2l + b + |m_u|$ and $p$. We expand the perpendicular acceleration $a_\perp$ (charge 1) and the axial acceleration
$a_\parallel$ (charge 0, real) in separate bases with total charge 1 and 0, and
$\ba_i = \operatorname{Re}(a_\perp)\bm e_1 + \operatorname{Im}(a_\perp)\bm e_2 +
a_\parallel \bm e_3$.

\paragraph{Purification.}
The pooled features are
\[
\tilde A_{h\nu A M_r L B M_u} = \sum_j f_c^h P_\nu(y) P_A(\hat r_\parallel)
P_L(2|\tilde u_\perp|^2 - 1) P_B(\tilde u_\parallel)\, \hat r_\perp^{M_r} \tilde u_\perp^{M_u},
\]
with linearisation as in 2D and the additional identities $|\hat r_\perp|^2 = 1 - \hat r_\parallel^2$ and
$|\tilde u_\perp|^2 = (\eta + 1)/2$, $\eta = 2|\tilde u_\perp|^2 - 1$, which keep each
factor a polynomial in a single variable.

\paragraph{Gravity.}
Agents moving under gravity are unlikely to respond symmetrically under
rotations that ignore gravity, so they can sensibly be expected to be symmetric
only under rotations about the vertical, since full rotation equivariance would
equate a neighbour above an agent with one beside it. In such a setting we use the fixed frame $\bm e_3 = \hat{\bm z}$
in place of the heading frame. The one-particle basis is unchanged and the centre
basis becomes
\begin{equation*}
  \psi_{pbm}(\bv) = P_p(2|\tilde v_\perp|^2 - 1)\, P_b(\tilde v_\parallel)\, \tilde v_\perp^{m}, \qquad q(\psi) = m.
\end{equation*}
The selection rule again requires the charges to add up to the
output charge, and purification is as before. The model is then exactly
equivariant under rotations about the vertical and is defined also for agents
at rest.

\paragraph{Reflections.}
A reflection through a plane that contains $\bm e_3$, with $\bm e_1$ chosen in
the plane, conjugates the perpendicular parts and leaves the axial parts
unchanged, so, as in 2D, real coefficients make both the heading-frame and the
vertical-frame model equivariant under it. Composed with the rotations, this
gives the heading-frame model equivariance under every reflection and the
vertical-frame model equivariance under reflections through vertical planes.

\subsection{Self-propulsion: the self term}
\label{sec:selfterm}

The memoryless model $\ba_i = \bm F(\bv_i, \mathcal{X}_i)$ describes an agent's
own motion only through its current velocity (the 1-body term $\bm F^{(1)}(\bv_i)$),
whereas in self-propulsion-driven agents, including the tracked fish of
Section~\ref{sec:realdata}, the own motion dominates the acceleration. dACE
therefore adds a self term, a function of the agent's own past.

The self term sees the agent's velocities $\bv_i(t - k\Delta)$ at a set of steps
$k \in K$ back, or the past accelerations $(\bv_i(t - (k-1)\Delta) - \bv_i(t -
k\Delta))/\Delta$ formed from them, or both, each called a channel $c$ with
scaled state $\tilde{\bm x}_c$ (with the input map of Section~\ref{sec:dace}). In 2D we describe every channel in the frame of the current heading $\hat v_i$
by $|\tilde x_c|$ and $w_c = \bar{\hat v}_i\, \tilde x_c$, both invariant
under rotations. An agent at rest keeps the heading of its latest non-zero
velocity, as in 3D, so $S$ is defined but not smooth at rest, and the self functions
are
\begin{equation*}
  S(\bv_i, \underbrace{\{\tilde{\bm x}_c\}_c}_{\text{past}}) = \hat v_i\, P_p(2|\tilde v_i| - 1) \prod_{c}
    P_{l_c}(2|\tilde x_c| - 1)\, w_c^{m_c},
\end{equation*}
with the product over a set of $T \in \{1, 2\}$ distinct channels and total degree
bounds for temporal orders $T = 1$ and $T = 2$. The factor $\hat v_i$ of charge 1 makes $S$ equivariant, and products over two
channels let the response to one past state depend on another, as in a glide that
follows a burst. In 3D we describe the channels by their perpendicular and axial parts in the frame of
the basis, as for neighbours. The self term is part of the
1-body term, and the interaction terms stay as before.

Since an agent's past motion is correlated with its neighbours', a self term fitted
jointly with the interactions risks absorbing part of them
(Section~\ref{sec:selfresults}). We therefore fit in stages: the interaction
functions alone first, then the self term on their residual, each with its own
regularisation. The interactions are then those of the fit without the
self term, and the self term models only what they leave unexplained. Our implementation also
provides the joint fit and the reverse order for comparison, and any other self
model, such as a neural network of the agent's own past, can replace the self term.

\subsection{Fitting and software}
\label{sec:fitting}

Coefficients minimise $\sum_{(\ba, \bv, \mathcal{X}) \in \mathcal{D}} |\ba - \bm F(\bv, \mathcal{X})|^2 + \lambda^2 \sum_{\psi\bm{k}} \Gamma_{\psi\bm{k}} |\tilde c_{\psi\bm{k}}|^2$
over the training set $\mathcal{D}$, where $\tilde c$ are the coefficients of the
column-normalised basis (in 3D separately for $a_\perp$ and $a_\parallel$), and
$\Gamma_{\psi\bm{k}} = 1 + \deg(\psi)^4 + \sum_t \deg(\phi_{k_t})^4$ for $B_{\psi\bm{k}} = \psi \prod_t
\phi_{k_t}$ is adapted from the algebraic smoothness prior of ACE potentials
\citep{witt2023acepotentials}, which penalises high polynomial degrees. The
regularisation $\lambda$ maximises the evidence (marginal likelihood) of the fit,
as in Bayesian ridge regression for ACE potentials. Unlike cross-validation, this
chooses $\lambda$ from the training fit alone, with no validation set.
We set the total degrees of the basis by hand, though this choice could be automated in
principle.
We solve the least-squares problem by a QR factorisation of the design matrix streamed
over chunks of environments \citep{golub2013}, so that memory does not grow
with the amount of data and the conditioning is that of the design matrix, not of
the normal equations. The solutions and the evidence for all $\lambda$ then follow
from one singular value decomposition of the triangular factor, one per stage in a
staged fit.

We implemented \texttt{dACE.jl} \citep{gutleb2026dacejl} in Julia \citep{bezanson2017},
building on the ACEsuit ecosystem \citep{witt2023acepotentials}. The one-particle bases come
from \texttt{Polynomials4ML.jl} \citep{polynomials4ml}, and the solvers of
\texttt{ACEfit.jl} \citep{acefit} are available for joint fits.

\section{Numerical experiments on synthetic data}
\label{sec:results}

Collectives have no first-principles reference, so an interaction law can only be
tested on simulated collectives, where the rule is known, or on tracked data, where
it is not. On simulated collectives we compare the learned law with the truth by the
relative RMS error $\|\hat{\bm F} - \bm F\| / \|\bm F\|$ over held-out environments
(the relative force error), by the same error of one body-order contribution, and in
free-running simulation. We score next-step predictions $\hat{\bm a}$ of observed accelerations $\bm a$ by $R^2 =
1 - \|\hat{\bm a} - \bm a\|^2 / \|\bm a\|^2$.

\subsection{Simulated collectives}
\label{sec:swarms}
We generate trajectories with Agents.jl \citep{datseris2022agents} and update all agents
synchronously by a semi-implicit Euler step, so that the finite
differences of the recorded data are exact samples of the rule.

All examples build on D'Orsogna-type self-propelled particles
\citep{dorsogna2006} with Cucker--Smale-type alignment \citep{cuckersmale2007}:
\begin{align}
  \ba_i &= \underbrace{(\alpha - \beta|\bv_i|^2)\bv_i}_{\text{1-body}}
  + \underbrace{\sum_j f_c(\rho_{ij}) \bigl[U'(\rho_{ij}) \hat\br_{ij} + \gamma\, e^{-\rho_{ij}/\ell_v} \bu_{ij}\bigr]}_{\text{2-body}}
  + \underbrace{\kappa_v \sum_{j<k} g_{ij} g_{ik}\, \tfrac{1 - \hat\br_{ij}\cdot\hat\br_{ik}}{2}\, \tfrac{\bu_{ij} + \bu_{ik}}{2}}_{\text{3-body}} \notag \\
  &\quad + \underbrace{\kappa_4 \sum_{j<k<l} g_{ij} g_{ik} g_{il} \Bigl(1 - \tfrac{|\hat\br_{ij} + \hat\br_{ik} + \hat\br_{il}|^2}{9}\Bigr) \tfrac{\bu_{ij} + \bu_{ik} + \bu_{il}}{3}}_{\text{4-body}},
  \label{eq:dorsogna}
\end{align}
with Morse potential $U(\rho) = C_r e^{-\rho/\ell_r} - C_a e^{-\rho/\ell_a}$ and
$g(\rho) = f_c(\rho) e^{-\rho/\ell_3}$. Each term is of exactly one body
order, so the true body-order decomposition is known. We consider four collectives:
\begin{description}
  \item[Pairwise] $\kappa_v = \kappa_4 = 0$: the classical setting, exactly of body order $\le 2$.
  \item[3-body flanking] $\kappa_v = 2$, $\kappa_4 = 0$: an agent flanked by two
    neighbours on opposite sides aligns with their mean velocity, a non-reducible
    three-body effect in the spirit of \citet{katz2011}.
  \item[Averaged alignment] we replace the alignment term by
    $\gamma \sum_j w_{ij} \bu_{ij} / (w_0 + \sum_j w_{ij})$, $w_{ij} = f_c e^{-\rho_{ij}/\ell_v}$,
    as in Vicsek-type and Motsch--Tadmor models \citep{vicsek1995,motschtadmor2011}. The
    normalisation makes the rule implicitly many-body at \emph{all} orders.
  \item[2+3+4-body] $\kappa_v = 2$, $\kappa_4 = 4$: adds a 4-body
    ``encirclement'' alignment, in which an agent aligns with any three
    neighbours, weighted by how evenly they surround it. All four body
    orders contribute with comparable magnitude.
\end{description}
To test dACE on rules outside this family, we add two classical models with
discontinuities that a smooth polynomial expansion cannot represent exactly:
\begin{description}
  \item[Boids] a simplified model in the spirit of \citet{reynolds1987} with averaged cohesion,
    alignment and separation, turning at a finite rate at constant speed, with a hard separation radius and a hard cutoff at $R$.
  \item[Zonal] a simplified zonal model in the spirit of \citet{couzin2002},
    without blind angle and noise: neighbours closer than 1 repel with priority, otherwise the agent aligns with neighbours at distance in $[1, 3)$ and is
    attracted by those in $[3, R)$. Its heading turns towards the resulting
    direction at a finite rate, at constant speed.
\end{description}
These two are implicitly many-body through averaging, normalisation and the
priority rule. Their agents start at the fixed speed at which both models move,
so that the data reflects interactions rather than the relaxation of initial
speeds.

\paragraph{Chiral rules.} All rules above are mirror symmetric. To test whether chirality can be learned, and
whether real coefficients then fail, we build chiral rules in 2D: we turn the 2-body
term of the pairwise collective by a fixed angle of
$10^\circ$ or $30^\circ$, add to it a constant turning $\omega(-v^y_i, v^x_i)$ with
$\omega = 0.3$, or weight the alignment term of the averaged-alignment collective by $1 + b \sin\varphi_{ij}$,
with $\varphi_{ij}$ the bearing of the neighbour, positive to the left, and $b = 0.3$
or $0.7$, so that agents follow neighbours on their left more strongly.

\paragraph{Stochastic self-propulsion.} To test models on data in which the
agents' own motion dominates, as for animals, we add to the pairwise and
2+3+4-body collectives an intrinsic drive $\bm d_i(t)$ in each agent's heading
frame, $\ba_i = \bm F(\bv_i, \mathcal{X}_i) + \bm d_i$. Each of its two components
has an RMS equal to that of $|\bm F|$ in a simulation without drive, so its mean
square is twice that of $\bm F$. The drive is either an Ornstein--Uhlenbeck process along
and across the heading with correlation time 0.5, or burst-and-coast swimming, a
thrust along the heading that pulses with a noisy phase of period 1, with the same
random turning.

\paragraph{Data.}
Each data point is one environment with the agent's finite-difference acceleration as
target, with a median of about ten neighbours within $R$. We train on \ntrainsims{} short simulations of 30 agents from random initial positions, headings and speeds, so that the data samples the
transient and not only the settled flock, test on 4 further simulations, and start
trajectory predictions from held-out initial conditions and run them to $t = 20$, well
beyond the training window. We report medians over \nsplits{}
independent splits (\nsplitsThreeD{} in 3D), with ranges in
Figure~\ref{fig:recovery}, and trajectory figures show the first split. In 3D the
agents start in a cube with headings uniform on the sphere, 40 of them so that
the number of neighbours matches 2D.

\paragraph{Models.} The dACE models use angular orders $|m_r|, |m_u| \le 1$ and
$l = 0$, in 3D also $a, b \le 1$ (so the one-particle functions are linear in the
velocities), and the same total degrees per body order for every collective, 6, 12, 8 and 6 for the
1- to 4-body terms in 2D and 6, 12, 7 and 5 in 3D, decreasing from the 2-body term on,
since the basis grows fastest with degree at high body order. We run SocialNet with the authors' code and training defaults (100 epochs with early stopping on a validation part of the training
samples, and as a check 1000 epochs). SocialNet has no radial cutoff and takes every other agent in the group as input. Its architecture is two-dimensional and
mirror symmetric, so we do not run it on the 3D benchmarks, and it cannot represent chiral rules well.

\subsection{Interaction laws body order by body order}

\begin{figure}[tbp]
  \centering
  \begin{subfigure}[b]{0.49\linewidth}\centering\includegraphics[width=\linewidth]{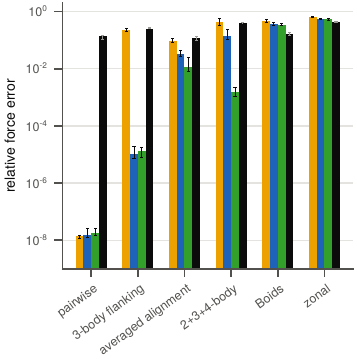}\caption{}\end{subfigure}\hfill
  \begin{subfigure}[b]{0.49\linewidth}\centering\includegraphics[width=\linewidth]{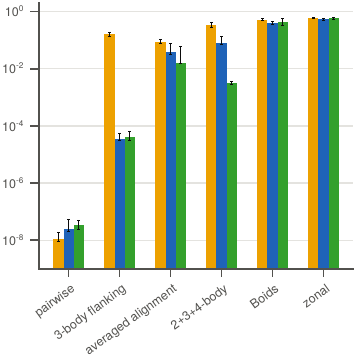}\caption{}\end{subfigure}
  \includegraphics[width=\linewidth]{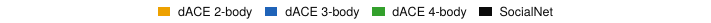}
  \caption{Relative force error of every model on every collective in (a) 2D, with
    SocialNet, and (b) 3D, median (bar) and range (whisker) over \nsplits{} splits in 2D
    (SocialNet 3) and \nsplitsThreeD{} in 3D. The error drops by orders of magnitude once
    the model's body order reaches the rule's, the 2-body model fails on every many-body
    rule, and the discontinuous Boids and zonal rules keep every model's error large.}
  \label{fig:recovery}
\end{figure}

To test whether dACE recovers a rule of known body order, and what each added body order
contributes, Figure~\ref{fig:recovery} shows the relative force error of every model on
every collective, in 2D with SocialNet and in 3D. On every smooth rule, dACE
truncated at the true body order recovers the law, the 2-body model fails as
soon as the rule has a many-body component, and SocialNet's errors stay far larger at either training length. For averaged alignment,
which has no finite body order, the error decreases with each added body order.
Boids and the zonal model are not smooth. Their discontinuities are
approximated only slowly by smooth polynomials, so dACE's error stays large at
every body order, also on the training data, and while SocialNet does slightly better on
these two, its error is also large.

To check that the fit assigns each contribution to the right body order,
Table~\ref{tab:recovery} gives the error of each learned body-order contribution against
the true term. On the pairwise collective the 4-body model adds almost no spurious 3- and 4-body terms,
on the 3-body collective almost no 4-body term, and on the 3-body and 2+3+4-body
collectives, the latter in 2D and 3D, every contribution matches its true term, so a
single linear fit separates the orders.

\begin{table}[tbp]
  \centering
  \caption{RMS error of each body-order contribution of the learned dACE 4-body model against
    the true term, relative to the RMS acceleration (median over splits).}
  \label{tab:recovery}
  \begin{tabular}{lllll}
    \toprule
    collective & 1-body & 2-body & 3-body & 4-body \\
    \midrule
    pairwise & \pairRecoveryZero{} & \pairRecoveryOne{} & \pairRecoveryTwo{} & \pairRecoveryThree{} \\
    3-body flanking & \recoveryZero{} & \recoveryOne{} & \recoveryTwo{} & \recoveryThree{} \\
    2+3+4-body & \mixedRecoveryZero{} & \mixedRecoveryOne{} & \mixedRecoveryTwo{} & \mixedRecoveryThree{} \\
    2+3+4-body, 3D & \mixedThreeDRecoveryZero{} & \mixedThreeDRecoveryOne{} & \mixedThreeDRecoveryTwo{} & \mixedThreeDRecoveryThree{} \\
    \bottomrule
  \end{tabular}
\end{table}

\begin{figure}[tbp]
  \centering
  \begin{subfigure}[b]{0.32\linewidth}\centering\includegraphics[width=\linewidth]{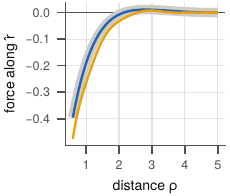}\caption{Radial pair force}\end{subfigure}\hfill
  \begin{subfigure}[b]{0.32\linewidth}\centering\includegraphics[width=\linewidth]{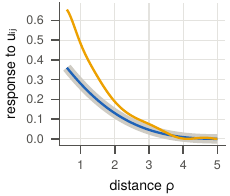}\caption{Pair alignment}\end{subfigure}\hfill
  \begin{subfigure}[b]{0.32\linewidth}\centering\includegraphics[width=\linewidth]{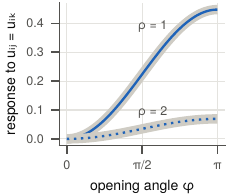}\caption{3-body flanking alignment}\end{subfigure}
  \includegraphics[width=\linewidth]{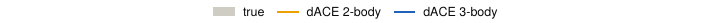}
  \caption{Interaction kernels learned from the 3-body collective: the learned 2-body
    contribution for one neighbour (a, b) and the 3-body contribution for two
    neighbours at distance $\rho = 1$ (solid) or 2 (dotted) and opening angle $\varphi$
    (c), true kernels as wide grey bands.}
  \label{fig:kernels}

\end{figure}

\begin{figure}[tbp]
  \centering
  \begin{subfigure}[b]{0.25\linewidth}\centering\includegraphics[width=\linewidth]{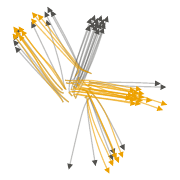}\caption{dACE 2-body}\end{subfigure}\hspace{0.05\linewidth}\begin{subfigure}[b]{0.25\linewidth}\centering\includegraphics[width=\linewidth]{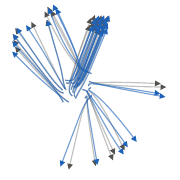}\caption{dACE 3-body}\end{subfigure}\hspace{0.05\linewidth}\begin{subfigure}[b]{0.25\linewidth}\centering\includegraphics[width=\linewidth]{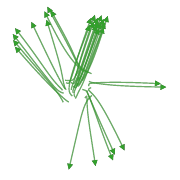}\caption{dACE 4-body}\end{subfigure}\par\begin{subfigure}[b]{1\linewidth}\centering\includegraphics[width=\linewidth]{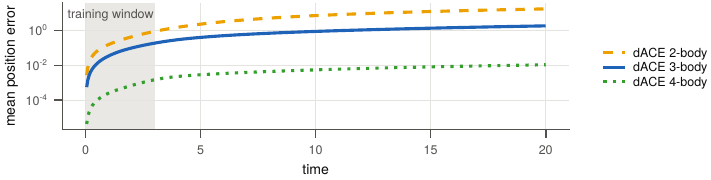}\caption{}\end{subfigure}
  \caption{Trajectory prediction for a held-out initial condition of the 2+3+4-body
    collective in 2D. (a--c) True (grey) and predicted paths: the 2-body and 3-body models
    drift away while the 4-body model stays close to the true paths. (d) Mean position
    error, the training time span shaded. This
    and other rollout animations are available on figshare \citep{gutleb2026figshare}.}
  \label{fig:rollout-mixed}

  \medskip

  \begin{subfigure}[b]{0.25\linewidth}\centering\includegraphics[width=\linewidth]{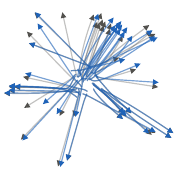}\caption{dACE 3-body}\end{subfigure}\hspace{0.05\linewidth}\begin{subfigure}[b]{0.25\linewidth}\centering\includegraphics[width=\linewidth]{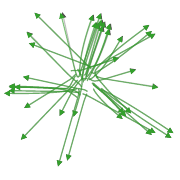}\caption{dACE 4-body}\end{subfigure}\par\begin{subfigure}[b]{1\linewidth}\centering\includegraphics[width=\linewidth]{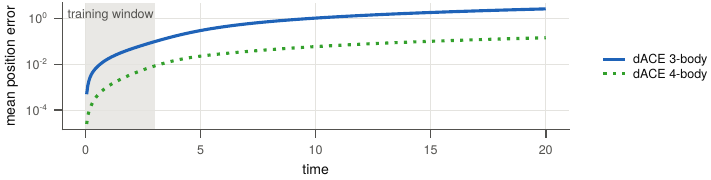}\caption{}\end{subfigure}
  \caption{As Figure~\ref{fig:rollout-mixed}, in 3D (40 agents) and without the 2-body model,
    whose rollout diverges, with paths projected onto the two principal axes of the true
    positions and errors computed in 3D.}
  \label{fig:rollout-mixed3d}
\end{figure}

To show that the learned law can be interpreted, Figure~\ref{fig:kernels} shows the
interaction kernels of the 3-body collective, obtained by probing the learned model with one or two neighbours,
and the distorted pair interaction by which the 2-body model absorbs part of
the 3-body effect. To test the learned laws beyond held-out environments, Figures~\ref{fig:rollout-mixed}
and \ref{fig:rollout-mixed3d} show trajectory predictions on the 2+3+4-body collective
in 2D and 3D.

\begin{figure}[tbp]
  \centering
  \includegraphics[width=0.6\linewidth]{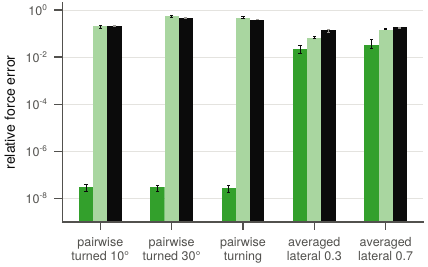}\\
  \includegraphics[width=0.6\linewidth]{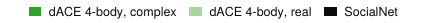}
  \caption{Relative force error on the chiral collectives for dACE 4-body with complex
    and with real (mirror-symmetric) coefficients and for SocialNet, median (bar) and
    range (whisker) over 3 splits. Complex coefficients learn the chiral rules, whereas
    real coefficients cannot represent chirality.}
  \label{fig:chiral}
\end{figure}

Figure~\ref{fig:chiral} shows that with complex coefficients dACE learns the chiral
collectives nearly as accurately as their mirror-symmetric versions.

\subsection{Data efficiency}
\label{sec:efficiency}

\begin{figure}[tbp]
  \centering
  \begin{subfigure}[b]{0.49\linewidth}\centering\includegraphics[width=\linewidth]{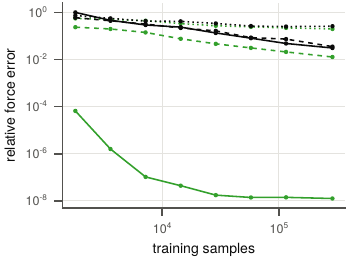}\caption{Pairwise}\end{subfigure}\hfill
  \begin{subfigure}[b]{0.49\linewidth}\centering\includegraphics[width=\linewidth]{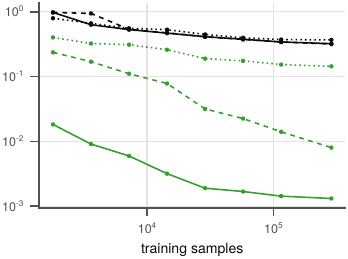}\caption{2+3+4-body}\end{subfigure}
  \includegraphics[width=\linewidth]{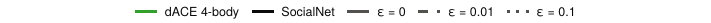}
  \caption{Relative force error on clean test simulations against the number of
    training samples, for dACE 4-body and SocialNet, with observation noise $\varepsilon$ (line style).}
  \label{fig:efficiency}
\end{figure}

To test how much data each model needs, and how noise changes that, we train dACE and SocialNet on
growing amounts of data with observation noise of size $\varepsilon$, which shifts every
recorded position and velocity by an independent vector of random direction and length
uniform in $[0, \varepsilon]$. dACE needs less data than SocialNet
(Figure~\ref{fig:efficiency}). On the 2+3+4-body rule, SocialNet improves only slowly
with data and hardly changes with the noise level, whereas dACE shows clear gains.

\subsection{Self-propulsion and next-step prediction}
\label{sec:selfresults}

To find out what next-step accuracy measures, we use the stochastic collectives (Section~\ref{sec:swarms}), where
the force $\bm F$ is known and the drive is time-correlated, so an agent's past predicts much of its next acceleration but
nothing about $\bm F$.

\begin{figure}[tbp]
  \centering
  \begin{subfigure}[b]{0.49\linewidth}\centering\includegraphics[width=\linewidth]{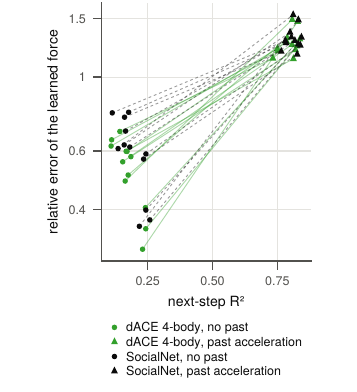}\caption{}\end{subfigure}\hfill
  \begin{subfigure}[b]{0.49\linewidth}\centering\includegraphics[width=\linewidth]{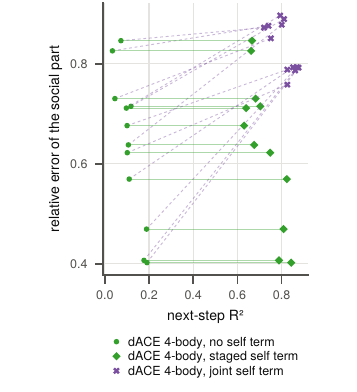}\caption{}\end{subfigure}
  \caption{Stochastic self-propulsion. Each point is one data set, a pairwise or 2+3+4-body
    collective with Ornstein--Uhlenbeck or burst-and-coast drive, in one of 3 splits. (a) The past
    acceleration raises next-step $R^2$ several-fold while the error of the learned force,
    for a model with past input its whole prediction, grows beyond that of predicting no
    force. (b) A self term raises $R^2$ in both fits, but the staged fit (horizontal lines)
    leaves the social part's error unchanged, whereas the joint fit (dashed) raises it.}
  \label{fig:selfpropulsion}
\end{figure}

Every model that uses the past, dACE with the past acceleration as an input or
SocialNet with its past input, raises next-step $R^2$ several-fold. Alarmingly, however, their
\emph{relative force errors grow}, beyond that of predicting no force at all (Figure~\ref{fig:selfpropulsion}a).
The gain in $R^2$ comes from the agent's own motion alone, which is correlated with its
neighbours' and can therefore stand in for part of the interactions (the reason we fit
the self term in stages). Next-step accuracy thus cannot rank interaction models. On
tracked data, where the law cannot be compared with the truth and models are usually
compared by held-out prediction accuracy \citep{heras2019}, this is a genuine difficulty with
no single solution. The best approach available is a combination of checks: prediction
under a protocol that shares no frames with the target, gain from adding
interactions to the self model, and consistency of the law between disjoint
sets of the data (Section~\ref{sec:realdata}).

A self term fitted jointly with the interactions behaves the same way. It raises $R^2$,
but its social part, the 2-body and higher terms, ends up further from the true force
than that of the memoryless fit. The staged fit keeps the memoryless interactions
and still reaches most of the joint $R^2$ (Figure~\ref{fig:selfpropulsion}b).

\section{Numerical experiments on tracking data}
\label{sec:realdata}

\begin{figure}[tbp]
  \centering
  \begin{subfigure}[b]{0.49\linewidth}\centering\includegraphics[width=\linewidth]{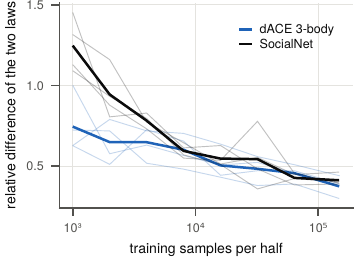}\caption{Zebrafish}\end{subfigure}\hfill
  \begin{subfigure}[b]{0.49\linewidth}\centering\includegraphics[width=\linewidth]{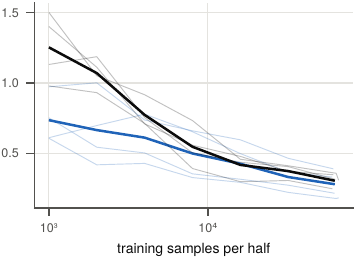}\caption{Golden shiners}\end{subfigure}
  \caption{Relative difference between the memoryless force laws learned from two disjoint
    halves of the tracked-fish data, for four held-out
    groups per species (faint) and their mean (bold).}
  \label{fig:realdata}
\end{figure}

\begin{figure}[tbp]
  \centering
  \includegraphics[width=\linewidth]{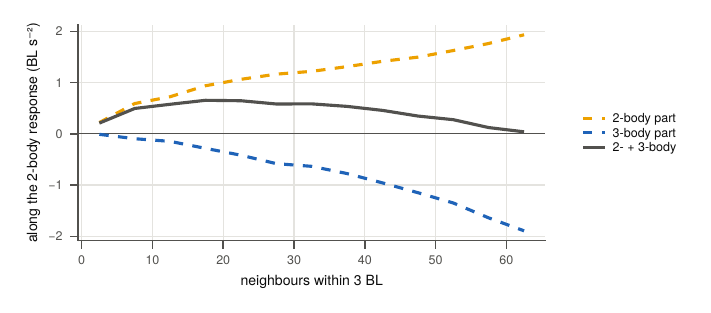}
  \caption{The 3-body term in the golden shiners. Mean component along the 2-body response
    of the learned 2-body part, the 3-body part and their sum, against the number of
    neighbours, at the configurations observed in the ten trials (dACE 3-body, cutoff 3\,BL).}
  \label{fig:shinersaturation}
\end{figure}

The zebrafish data we explore in this section is the eight of the reported nine videos of groups
of 60 to 100 fish of \citet{heras2019} whose validated idtracker.ai tracking
\citep{torrents2026} is publicly released, with the 2019 tracking of the same videos
\citep{heras2019,romeroferrero2019} as a second data set. The golden shiner data is ten trials with groups of 10 to 151 fish from the experiments
of \citet{katz2011}, re-tracked by \citet{davidson2021}. We
use steps of about 0.1\,s, lengths in body lengths (BL), targets away from the
walls, and neighbours within 8\,BL for zebrafish and 3\,BL for shiners. We fit dACE 3-body with the saturating input map \eqref{eq:speedmap} and a staged self term,
with total degrees 32, 12 and 6 for the zebrafish and 4, 6 and 4 for the shiners, angular orders $|m_r|, |m_u| \le 4$ and $l \le 2$,
leaving one video or trial out for testing. Inputs come from frames up to $t$ only and targets from frame $t + 1 + G$ onward,
with $G \in \{0, 3\}$ skipped frames, so that a past input shares no frame, and with it no tracking
error, with the target.
All fits use real coefficients. Complex coefficients allow chiral rules but improved
nothing for either species, which suggests that these fish have no preferred side.
We check the learned laws in three ways: the gain in prediction from adding interactions to the self model (Table~\ref{tab:gain}), the consistency of the law between disjoint halves of the data (Figure~\ref{fig:realdata}), and for the shiners the learned 3-body term itself (Figure~\ref{fig:shinersaturation}).

Table~\ref{tab:gain} gives the $R^2$ of the self model alone and with interactions.
Adding interactions improves prediction in both species, at both $G$, for every zebrafish video
and every shiner trial, and the 3-body model beats the 2-body model on every shiner trial.
The self term explains far more of the acceleration than the interactions.
SocialNet extracts a larger social gain from the zebrafish than dACE, but from a far weaker self
model. A larger gain does not mean better interactions (Section~\ref{sec:selfresults}), since
whatever the self model misses of the agent's own motion can be absorbed by the interactions, and
with a past input that shares no frame with the target SocialNet predicts worse than dACE on all
eight videos.

\begin{table}[tbp]
  \centering
  \caption{$R^2$ in percent of the self model alone and with interactions, pooled over the videos or
    trials left out of the fit in turn, and the gain $\Delta$ in percentage points, for $G$ skipped frames.
    SocialNet's past input is lagged so that it shares no frame with the target.}
  \label{tab:gain}
  \begin{tabular}{ll@{\hspace{1em}}rrr@{\hspace{1em}}rrr}
\toprule
& & \multicolumn{3}{c}{$G = 0$} & \multicolumn{3}{c}{$G = 3$} \\
\cmidrule(lr){3-5} \cmidrule(lr){6-8}
data & model & self & with & $\Delta$ & self & with & $\Delta$ \\
\midrule
zebrafish & dACE 3-body & 10.5 & 11.9 & 1.4 & 3.8 & 4.9 & 1.1 \\
& SocialNet, no past input & 1.2 & 3.1 & 1.9 & 1.5 & 3.4 & 1.9 \\
& SocialNet, past input & 5.6 & 8.7 & 3.1 & 2.4 & 4.2 & 1.8 \\
golden shiners & dACE 3-body & 9.4 & 11.0 & 1.6 & 5.7 & 7.1 & 1.3 \\
& SocialNet, past input & 4.0 & 7.7 & 3.7 & 3.1 & 4.9 & 1.7 \\
\bottomrule
\end{tabular}

\end{table}

Laws learned from disjoint halves of the data that disagree cannot both be correct, so
we measure the consistency of a learned law as follows. We fit the memoryless force law
of a model, with no past input and no self term, to two disjoint halves of the data,
alternating blocks of time, dropping the samples whose frames reach across a block
boundary, and measure the disagreement of the two learned laws $\hat{\bm F}_A$ and
$\hat{\bm F}_B$, 1-body term included, on the same held-out environments, $\|\hat{\bm F}_A - \hat{\bm F}_B\| / ((\|\hat{\bm F}_A\|^2 +
\|\hat{\bm F}_B\|^2)/2)^{1/2}$ (Figure~\ref{fig:realdata}). Consistency is meaningful only if neither model is regularised into agreement. Both
halves therefore get the same regularisation, for dACE the $\lambda$ that maximises
their summed evidence and for SocialNet the training length with the lowest summed
validation error, and the measure is relative, so that a uniform shrinkage does not
improve it.

In our experiments, dACE's law is generally more consistent than SocialNet's (Figure~\ref{fig:realdata}). With 1000
samples per half, SocialNet's two laws disagree by more than their own size in both species,
far more than dACE's. Regularising each half on
its own changes neither model much.

For the shiners, Figure~\ref{fig:shinersaturation} shows the effect of the
3-body term by plotting the 2-body and 3-body parts of the fitted dACE model against
the number of neighbours. In sparse neighbourhoods the 3-body term is small. The 2-body sum grows with the number of
neighbours, whereas the 3-body term opposes it more and more, so the full response
stops growing. With each trial held out of the fit in turn, the 3-body term opposes
the 2-body response on every trial with dense neighbourhoods. This agrees with \citet{katz2011}, who found three-body forces in groups of
three and weaker longer-range forces in larger groups.

\section{Discussion}
\label{sec:limitations}

dACE learns interaction laws body order by body order in 2D or 3D, and with less
data than competing models, and on synthetic and tracked data alike the learned laws
can be read off and interpreted. Beyond dACE itself, we draw two broader conclusions. First, data can decide whether pairwise interactions suffice. The learned higher-order contributions
are negligible where the rule is pairwise and large where it is not, and in the
shiners the response to many neighbours is not the sum of the responses to one. Other tracked groups, such as the
jackdaw flocks of \citet{ling2019}, the surf scoter flocks of
\citet{lukeman2010} and the midge swarms of \citet{sinhuber2019}, are open to
the same tests.
Second, next-step accuracy is a poor guide
to interaction models. In our simulations it rises while the learned interactions
get worse. No single measure is safe on tracked data, so we recommend judging interaction laws by several checks together, such as prediction under a protocol that
shares no frames, the gain from adding interactions, and the consistency between
disjoint sets or rolled-out trajectories. Several future research directions remain open:
\begin{description}
  \item[Bounded influence.] Atoms respond to forces by their nature. Animals act on
    perceived information through a limited capacity, so the influence of their
    neighbours is bounded, by selecting a few of them (starlings appear to interact with their
    nearest six or seven \citep{ballerini2008}), by occlusion \citep{davidson2021},
    or by saturation, as in the shiner data discussed above. A fixed metric neighbourhood captures
    saturation, but selection and occlusion change the shape of the neighbourhood itself. A
    $k$-nearest or visibility-based neighbourhood, with the body-ordered expansion on
    top, would let these mechanisms be compared directly and thus learned from data.
  \item[Walls, obstacles and other species.] Obstacles could enter as a static second
    species of particles, as ACE handles chemical elements, and so could predators
    or the members of mixed-species groups. Free-running tests of tracked groups need
    such boundaries.
  \item[Discontinuous rules.] Rules with hard zones or priority switching are
    approximated slowly by smooth polynomials, as seen in the Boids and zonal experiments.
    Piecewise polynomial one-particle bases are a natural candidate to fix this and are compatible
    with purification, but their knots must be known in advance or learned.
\end{description}

\ifjournal
\section*{Ethics}
This work did not require ethical approval from a human subject or animal welfare
committee.
\fi

\section*{Data accessibility}
\texttt{dACE.jl} and the scripts that generate every figure, table and number
of this paper are available under the MIT licence \citep{gutleb2026dacejl}. The paper uses the zebrafish
trajectories of \citet{heras2019} in the trackings of \citet{romeroferrero2019}
and \citet{torrents2026} and the golden shiner trajectories of
\citet{davidson2021}, all openly available from their original publications.

\section*{Declaration of AI use}
We used language models, in particular Claude Opus 5.5 and Fable 5.1 and GPT-6 Luna and
Astra, in an assistive role to help with prototyping, to improve readability and in assistant and antagonistic reviewer roles. All ideas, experiments, analysis and conclusions originate with the author who takes full ownership and responsibility for this work.

\ifjournal
\section*{Authors' contributions}
T.S.G.: conceptualization, data curation, formal analysis, investigation,
methodology, software, validation, visualization, writing---original draft,
writing---review and editing.

\section*{Conflict of interest declaration}
We declare we have no competing interests.
\fi

\section*{Funding}
This work received no specific funding.

\section*{Acknowledgements}
We thank Leah Edelstein-Keshet, Ryan Lukeman, Cheuk Hin Ho and Christoph Ortner
for insightful early discussions about the project.

\end{document}